# "I Wrote, I Paused, I Rewrote" — Teaching LLMs to Read Between the Lines of Student Writing

Samra Zafar*
Shaheer Minhas*
samrazafar003@gmail.com
minhasshaheer0@gmail.com
FAST National University of
Computer and Emerging Sciences
Lahore, Punjab, Pakistan

Arfa Naeem
Zahra Ali[†]
edu.mattrs@gmail.com
zahra.ali1801@gmail.com
FAST National University of
Computer and Emerging Sciences
Lahore, Punjab, Pakistan

Syed Ali Hassan Zaidi[‡]
sali.zaidi572@gmail.com
FAST National University of
Computer and Emerging Sciences
Lahore, Punjab, Pakistan

## Abstract
Large language models (LLMs) like Gemini are becoming common tools for supporting student writing. But most of their feedback is based only on the final essay—missing important context about how that text was written. In this paper, we explore whether using writing process data, collected through keystroke logging and periodic snapshots, can help LLMs give feedback that better reflects how learners think and revise while writing. We built a digital writing tool that captures both what students type and how their essays evolve over time. Twenty students used this tool to write timed essays, which were then evaluated in two ways: (i) LLM generated feedback using both the final essay and the full writing trace, and
(ii) After the task, students completed surveys about how useful and relatable they found the feedback. Early results show that learn- ers preferred the process-aware LLM feedback, finding it more in tune with their own thinking. We also found that certain types of edits, like adding new content or reorganizing paragraphs, aligned closely with higher scores in areas like coherence and elaboration. Our findings suggest that making LLMs more aware of the writing process can lead to feedback that feels more meaningful, personal, and supportive.

*Both authors contributed equally to this research.

**ACM Reference Format:**
Samra Zafar, Shaheer Minhas, Arfa Naeem, Zahra Ali, and Syed Ali Hassan Zaidi. 2018. "I Wrote, I Paused, I Rewrote" — Teaching LLMs to Read Between the Lines of Student Writing. In *Proceedings of Make sure to enter the correct conference title from your rights confirmation email (Conference acronym 'XX)*. ACM, New York, NY, USA, 7 pages. https://doi.org/XXXXXXX.XXXXXXX

## 1 Introduction
In schools, Large Language Models (LLMs) are increasingly used to provide feedback on student writing. They're easy to get, under- stood quickly and usually helpful, though a problem can occur. Most software today responds only to the final draft of a text and gives feedback that is static and general. In turn, this lets educators forget that students didn't choose their issues; the challenges affected their progress along the way. The task of writing means constantly stop- ping, starting again, fixing up the work and dealing with moments you're not sure about. If feedback doesn't focus on the student's past experience, it may just provide general tips—instead of helping them grow in learning [Chan et al. 2024; Jansen et al. 2025].

Much of the research community now supports the idea that good feedback covers the whole process, not just the final outcome. It should represent how students actually form their arguments. As illustrated by Schiller et al. [Schiller et al. 2024], static systems don't see that there might be varying student engagement depending on the stage of writing—being it the initial planning or mid-writing stage. Even though using LLMs increases interest in the subject [Chan et al. 2024], their lack of thought-processing capabilities de- creases what they can offer in education. This is important because taking breaks, preparing work and changing your plan are big parts of becoming a strong writer [Fagbohun et al. 2024; Lee et al. 2024]. To help with understanding and supporting those skills, a few re- searchers are using things like records of keystrokes and frequently captured drafts by students. They make it possible to observe stu- dents' writing as it happens, seeing what they write and how they approach it. With ScholaWrite [Wang et al. 2025], more details are collected at each moment to display how students prepare, change and think about their written work. Another study done by Meyer and Jansen [Meyer et al. 2025] found that both early performance and motivation affected the way students react to feedback, imply- ing that context-specific, personalized systems could be valuable. For this work, we want to direct LLM-based feedback on writing toward a model guided by the real-life process of writing. We plan to create systems by combining keystroke data and timed snap- shots that are able to determine how students think, not only what they put down on paper. By observing students as they come, you can give help when they truly have a difficulty, instead of giving feedback once they are outside the classroom.

The main things we add to the field are: RQ1 (Cognitive Grounding): To what extent does access to writing process data (keylogging
+ snapshots) improve the LLM understanding of the cognitive process behind student essays? RQ2 (User Experience Trust): How do learners perceive the feedback quality and transparency when the LLM uses their writing process in its evaluation?

## 2 Related Work

Even though LLMs can now produce more fluent and useful feedback, they still fail to offer enough help with enhancing writing skills over time. Even though LLMs follow guidelines like the Common European Framework of Reference (CEFR), they don't recognize how students learn as they write, according to Benedetto et al. [2025].

More effort is being put into using behavioral and process data to close this difference. According to Schiller et al. [2024], students do not use feedback steadily; their level of engagement with it changes during the entire writing assignment. These behaviors which include lots of editing or long pauses, hint at the level of mental engagement a student has at each point. Regrettably, LLM-based programs continue to view writing as unchanging and do not consider these signs at all.

Some positive changes are taking place. Thanks to the ScholaWrite dataset Wang et al. [2025], a clearer picture is given of how ideas are developed and altered. Now it's more possible to understand design from different angles other than only the end result. Lee et al. [2024] state that working with students on writing support should start while they plan or draft their work, not only once things are already written.

Visually, students often show up with three potential thought bubbles, because they don't respond identically to feedback. It was discovered by Meyer and Jansen Meyer et al. [2025] that students' reactions to feedback depended on both their motivation and how early in their education they had success. So, we need to make sure feedback changes both in what is said and in its delivery method. Even so, few systems are able to match writers' real-time input with feedback from LLMs to accurately show students how they write. Research plays a key role here. We hope that by detecting how students pause, restructure and rewrite, the feedback system can give mentoring-level help instead of only saying basic things about the writing.

## 3 Methods
### 3.1 Text Collection Interface

For the collection of human written essays and their respective keylogged data, we designed our own interface using Javascript for the frontend and Python for the backend processing. Our interface uses javascript because of two main reasons: Javascript is a widely adopted language which can run on any web browser and is easy to deploy. Secondly, javascript makes it easier to log user behaviours like mouse clicks and keystrokes which were the core functionality of our custom interface [1]. The interface displays a randomly selected topic, out of a total set of 15 'generic' essay topics belonging to 3 categories: argumentative, reflective and analytical. We employed a well thought keylogging design where logs were generated on two events: whenever the user presses backspace key, the old log containing all the typed content so far, is saved. A new log begins on the backspace release. This design makes sure to save whatever content the user has typed before pressing the backspace key. Additionally, our design also ensures that it starts a new log, only after the user has released the backspace key for at least 3 seconds, ensuring that consecutive backspaces do not generate re- dundant logs and waste computational resources. Timestamps were saved with each log as well.

### 3.2 LLM Feedback Generation and Cognitive Grounding

The interface also captures the textbox snapshots every 3 minutes (6 snapshots for the total duration of the essay) in order to capture the full content context in addition to the fine grained typing behaviour captured through keylogging [1]. Our LLM API is hit upon every snapshot or keystroke log event, to keep the LLM active and context aware in real time. The system transmits the current state of the essay, along with metadata such as timestamps and revision actions. Once the user presses the submit button, the data including the final essay, the snapshots as well as the keystroke logging data are passed to the backend LLM API. We used Gemini API for this study because of the affordability and a generous token generation limit. The LLM generates a two fold feedback: part one contains the feedback along 4 rubrics against which the essay is evaluated including Thesis and Arguments, Language Use, Prompt Relevance and Organization/Structure of the essay. Part two of the feedback targets only the revision behaviour of the user. The LLM discusses the writing and pausing/revising process of the user and connects it with explicit timestamps or time-tagged snapshots as well. Following are a few examples of the LLM revision feedbacks, showing the time and context awareness of the LLM:

*"A significant number of backspaces and revisions between the 15:56–16:00 timestamp reflect a moment of decision-making around the thematic focus, particularly between the terms "competence" and "persistence," with "persistence" ultimately chosen as the guiding theme…"*

*"Snapshots captured between the 3–6 minute mark reveal the gradual development of main arguments regarding part-time jobs. Nonetheless, a prolonged pause followed by substantial rewriting near the conclusion suggests difficulty in articulating a strong closing statement."*

*"A marked pause around the 15-minute point precedes the addition of content related to positive outcomes, indicating a potential strategic shift in the student's argumentative framing."*

### 3.3 Data Collection

We conducted the study in an educational environment, targeting undergraduate students who are not native English speakers but have their undergraduate studies in English medium. We recruited 20 participants for this study through online academic channels and voluntary sign-ups. Participants completed a 20 minute essay writ- ing task. All the collected data was anonymized and consented for ethical reasons. A guidelines page was shown to each user stating about the duration, writing instructions and data usage guidelines (including keylogging and anonymity of data). After the submission of the essay, the LLM generated feedback, based on the actively logged keystroke and snapshot data, was shown to the user. The user was prompted to read the feedback thoroughly and then redirected to a short survey. The survey consisted of six 5-point Likert scale questions and 4 open ended questions (one of which was about the suggestions for improvement in our custom module).

**Table 1: Survey Questions and Sample Responses**

| Survey Question Codes | Sample Participant Responses | Thematic |
|---|---|---|
| 1. Which part of the feedback felt most accurate or helpful? | "It highlighted the right issues… my thesis was unclear and the argument lacked coherence." | Capturing Core Writing Issues |
| | "The summary of revisions was spot on; it reflected where I struggled the most with grammar and flow." | Capturing Core Writing Issues |
| 2. Was there any part that misunderstood your thinking or intention? | "It pointed out errors that I had already fixed, which felt unfair and frustrating." | Precision and Fairness in Feedback |
| | "The system didn't realize why I removed some sentences; it assumed I was avoiding complexity." | Precision and Fairness in Feedback |
| 3. Do you think the system understood why you revised certain parts of your essay? | "Yes, it tracked my struggle with word choice and helped me improve clarity." | Understanding the Revision Process |
| | "Mostly yes, but sometimes it missed when I made small surface edits rather than substantive changes." | Understanding the Revision Process |
| 4. What would you like the system to do better in future feedback? | "I want more personalized examples of vocabulary and informal language options to make writing livelier." | Aspirations for Personalization and Growth Support |
| | "It should understand emotional tone better and provide feedback that fits my writing style." | Aspirations for Personalization and Growth Support |

ChatGPT-4o helped us conduct the thematic analysis of student replies to the post-essay task survey for 18 respondents. Though thematic analysis always meant a lot of effort on the part of researchers [Braun and Clarke 2006], new LLMs are now making it easier and more reliable for qualitative research. Using GPT-3.5 as an example, [Dai et al. 2023] described an approach where human coders work with a LLM using ICL prompts to handle all tasks: code generation, reviewing codes, creating a set of codes and concluding with the main theme. Based on their findings, a person working with a coding model can achieve as much accuracy as or better than, two human coders by making the process faster and more properly connected.

It is also supported by [Qiao et al. 2025] which demonstrated that LLMs are useful in processing semi-structured interviews from areas such as psychology and public health. Likewise, [Tai et al. 2024] mentioned that using LLMs can improve review tasks in qualitative studies by marking key themes in responses that contain complex points. Yet, [Han et al. 2025] pointed out that LLMs are not as good at more in-depth context analysis which highlights the role of human involvement, as seen in the system suggested by [Dai et al. 2023].

Our team began like this: humans prepared some exemplars and became familiar with the data and after that, the LLM discovered the recurring categories and patterns in the data. The initial codes were constantly updated with help from the human researcher and the LM as a result of detailed negotiation, following the procedure proposed by Braun and Clarke (2006). Through teamwork, the themes found by the model fit the project needs and also relieved some effort from researchers. According to Dai et al., we also saw that the LLM created reliable code when given guidance and worked with explanations from expert models.

## 4 Results and Discussion

The main themes discovered by thematic analysis of 18 student feedback included five different topics. Feedback provides insights
into where the tool succeeded, where improvements are required, and what aspects of the LLM feedback were not agreeable by the participants.
At the start, the authors review 47 initial codes that come from students' responses about AI in their writing feedback. After studying raw codes, the authors summarized all the information into five crucial themes. Through this consolidation, the diversity of the data is retained and there is a better way to display and analyze the research according to its main goals.

Theme 1: Capturing Core Writing Issues
One of the clearest patterns, mentioned by a striking 15 out of 18 participants (83%), was how well the tool pinpointed central challenges in their writing. When asked, "Which part of the feedback felt most accurate or helpful?", many pointed to categories like Thesis Argumentation, Language Use, and the Revision Summary. These areas seemed to zero in on well-known weak spots like unclear thesis statements, disorganized structure, or awkward grammar. As one participant put it, "It highlighted the right issues… difficulty in expressing complex ideas, lack of coherence…" That line echoed a common sentiment: the tool was surprisingly good at catching those moments of struggle that writers often sense but can't always name. In that way, it seemed to work less like a grammar-checker and more like a writing coach who knows where students tend to get stuck.

Theme 2: Precision vs. Fairness in Feedback
While many students appreciated how detailed and thorough the feedback was, 56% felt uneasy about how their earlier mistakes were still being highlighted (even after they'd been fixed). When asked, "Was there any part that misunderstood your thinking or intention?", several participants pointed out that the system did notice their changes from draft to draft, but still seemed to dwell on those initial errors a bit too much. As one person shared, "It discusses mistakes that were then corrected… still the mention of them in the feedback seems a little critical." As a result, the system is aware of changes happening over time, though it should look for smoother solutions to prevent being seen as too critical. If the platform points out positives as well as negatives, it could increase users' faith in its fairness.

Theme 3: Tracking the Revision Journey
Interestingly, 13 participants (72%) felt the tool actually did a decent job tracking why they made certain changes, especially in areas tied to clarity, vocabulary, and idea development. When asked "Do you think the system understood why you revised certain parts of your essay?", one response stood out: "The system understood where I struggled to form ideas… later it was about improving word choice." Comments like these suggest the AI wasn't just reacting to surface errors—it was following a thread, noticing where the writer wrestled with getting their point across and where later changes reflected deeper clarity. That said, there were still instances where users felt it lumped minor edits in with bigger conceptual changes. The distinction between fixing a typo and reworking a paragraph wasn't always clear in the feedback, which left some users wishing for a more nuanced reading of their revision process.

Table 2: Thematic Summary of Participant Feedback Responses

| Theme Number | Theme Name | Description | Sample Codes | Response Coverage (n=18) |
|---|---|---|---|---|
| 1 | Capturing Core Writing Issues | Identification of thesis clarity, grammar, organization, and writing weaknesses | Thesis clarity, Argument strength, Language mechanics, Revision summary | 15 (83%) |
| 2 | Precision and Fairness in Feedback | Fairness and accuracy concerns regarding flagged errors, especially in revised drafts | Over-flagging, Unfair criticism, Temporal confusion | 10 (56%) |
| 3 | Understanding the Revision Process | AI's ability to model why and how students revise their writing, tracking cognitive strategies | Revision rationale, Word choice struggle, Idea formation | 13 (72%) |
| 4 | Perceived Gaps in Feedback Interpretation | Feedback rigidity, lack of genre/context sensitivity, misinterpretation of creative or concise writing | Bias detection, Detour labeling, Depth expectations | 8 (44%) |
| 5 | Aspirations for Personalization and Growth Support | Desire for adaptive, personalized feedback with vocabulary examples and psychological insight | Personalized rubrics, Informal language inclusion, Psychological profiling | 12 (67%) |

Theme 4: Missing the Nuance
Despite these strengths, 8 out of 18 participants (44%) mentioned moments where the system's interpretations didn't quite land. Specifically, responses to more creative or emotionally nuanced writing often felt misread. For instance, in reply to a prompt, one user explained, "The system considered my inflation example a detour, which I think was not." Another was frustrated by what they felt was rigid assessment within a strict word limit, noting, "Expected more complexity and depth even within the 250-character limit." These examples reveal a deeper challenge: the system sometimes missed the rhetorical choices writers were making intentionally, whether it was a matter of tone, genre, or just smart brevity. Writing, after all, isn't always about saying more; sometimes it's about saying just enough. Here, the feedback was not able to see the value in that flexibility. This agrees with Jansen and Meyer's research (2025) which explains that since students respond to writing activities in their own ways, feedback has to consider the mental process behind their writing in addition to checking the content.

Theme 5: Hoping for Personal Growth and Tailored Support
Lastly, 12 participants (67%) expressed a desire for feedback that felt more personal and growth-oriented. This included calls for more examples, clearer rubrics, and support for different writing styles—especially informal or creative approaches. One participant summed it up bluntly: "Add informal language… without it, essays would be boring." Another added, "Provide vocab examples to help students understand what kind of words they can use." Interestingly, while the feedback under Theme 5 was rich and passionate, much of it extended beyond the immediate scope of our research questions. They also highlight a slight shift away from our core investigation, which is specifically about how LLMs interpret revision behavior and context across drafts. In short, their feedback was valuable and eye-opening, even if it wasn't always directly tied to our central inquiry.

What stood out in the LLM-based system is that it could follow how changes happened over time and notice recurring patterns in what people revised. A lot of people mentioned that the tool knew where and when they struggled and understood the purpose of their edits, mainly in terms of choosing words, how everything was structured and forming their thesis. Having temporal and contextual sensitivity is an important improvement in how computers give feedback on writing. Even so, some expressed their wish to deepen their understanding of what they called deep running arguments. They believed the system should be aware of changes and they wanted it to also know why those changes were made, especially for tone, genre or creativity. In short, the LLM performed well at understanding the structure of texts but users preferred it to be more adaptable and sensitive to different ways of expressing ideas. The differences between how behaviors are tracked and interpreted point out both the strengths and weaknesses of today's AI technologies.

## 4.1 Quantitative Analysis

Responses from 18 participants on a five-part Likert scale explain that perceived usefulness and accuracy of the AI feedback tool are on a mostly positive trend. Users gave high scores to the ratings, reporting that that the system usually found the thesis and argument of the reviews correctly, with an average of about 4.1 out of 5. Respondents consistently awarded the highest scores (approximately 4.7) on the question "The language and grammar feedback was relevant and helpful," indicating that language and grammar advice was appreciated. Although participants really valued seeing how well the system followed their hesitant and revising moments, the scores for these aspects were lower, at 4.2 and 3.9, respectively. It means that the model is seen as accurate by many experts but not everyone agrees. In particular, users reported that the feedback they received was very helpful for developing their writing (mean 4.6), suggesting that the tool is seen as valuable for improvement. Respondents gave comments about how engaged the prompts were and the overall score was very close to 4.4. The numbers and the interviews support the idea that the system gives good language feedback, but it could do better at helping students think deeply and about fairness.

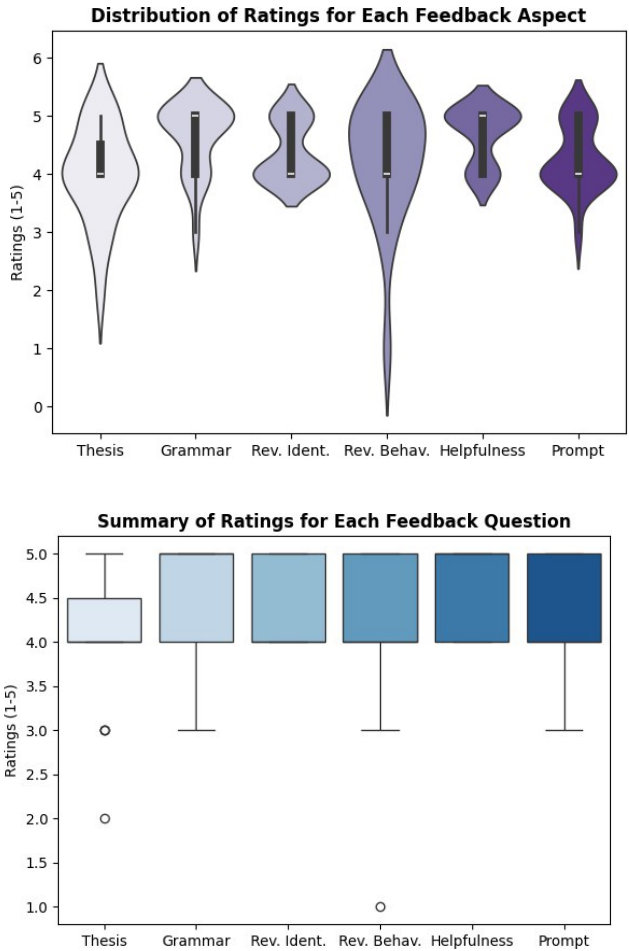

**Figure 1: Top: Violin plot showing distribution of participant responses across six Likert-scale items assessing perceived usefulness, relevance, and interpretability of AI-generated feedback. Bottom: Box plot depicting median, interquartile range, and outliers for each rating item, illustrating central trends and variation.**

## 4.2 Triangulation of Findings

In analyzing participants' perceptions of the LLM-generated feedback, several interrelated themes emerged, reflecting a nuanced and generally positive reception of the system's performance. The most prominent theme, Capturing Core Writing Issues, appeared in 83% of qualitative responses. Students consistently praised the sys- tem's ability to identify foundational elements such as thesis clarity, argument structure, and grammar. This finding was strongly cor- roborated by high Likert-scale ratings for items assessing thesis and structural alignment (M = 4.1) and the relevance of grammar-related feedback (M = 4.7), suggesting a robust convergence between sub- jective impressions and quantitative measures.

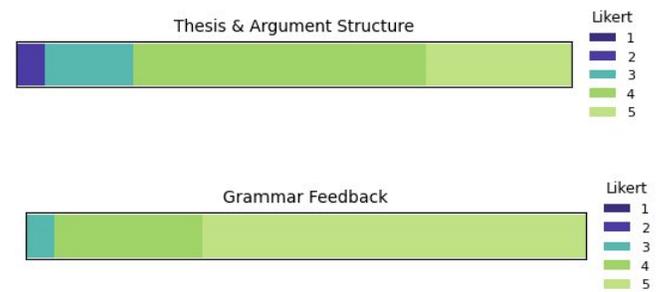

**Figure 2: Students' Perception of Feedback Accuracy on The- sis and Argument Structure(Top) and Grammar(Bottom)**

The theme Precision and Fairness in Feedback was also salient, appearing in 56% of qualitative responses. While participants generally appreciated the system's time-sensitive awareness, several voiced concerns about being judged for errors they had already revised, indicating a perceived lack of contextual grace. This nuance was not directly captured in the Likert scale, though it was reflected in relatively lower variance within the "engagement with the prompt" item (M = 4.4). The data suggest that although students accepted the system's assessments, they desired a more balanced evaluative tone that accounted for their revision history. A third

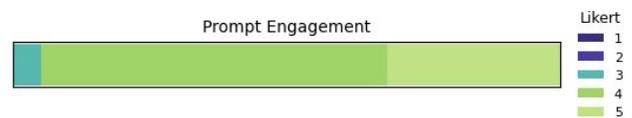

**Figure 3: Perceived Fairness and Insightfulness of Feedback on Prompt Engagement**

theme, Understanding the Revision Process, was present in 72% of responses and reflected positively on the system's capacity to track and interpret revision behaviors. Many students felt that their edits—often made in response to cognitive difficulty, organizational concerns, or moments of uncertainty—were meaningfully understood. This sentiment aligned with Likert items assessing whether the system recognized revisions and accurately modeled thought processes (M = 4.2 and M = 3.9, respectively), indicating that the tool offered a credible representation of students' metacognitive activity during writing. However, a notable minority (44%) pointed

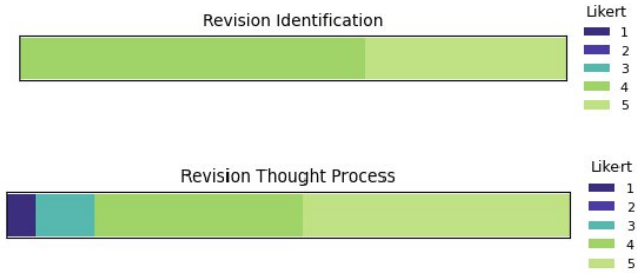

**Figure 4: Students' View on the System's Ability to Identify Points of Revision or Hesitation (Top) and Alignment Be- tween Feedback and Students' Perceived Revision Thought Process (Bottom)**

to Perceived Gaps in Feedback Interpretation, particularly when their responses included emotional nuance, creative expression, or constraints such as strict word limits. While this misalignment was expressed clearly in open-ended responses, it was not directly addressed by any Likert item. Nonetheless, lower outlier scores (ranging from 2 to 3) on items related to thesis clarity and prompt engagement may point to this interpretive dissonance. This limitation illustrates the challenges of using standard survey instruments to capture rhetorical or contextual nuance in LLM feedback evaluation.

Finally, Aspirations for Personalization and Growth Support were voiced by 67% of students, many of whom expressed a forward- looking desire for more adaptive, human-like interactions—ranging from tone-sensitive suggestions to vocabulary enrichment and psy- chologically grounded interpretations. Interestingly, this theme coexisted with the highest overall Likert mean across all items: "The feedback was helpful for improving my writing skills" (M = 4.6). This suggests that while the system met current expectations, students are eager for its evolution into a more personalized and context-aware feedback partner.

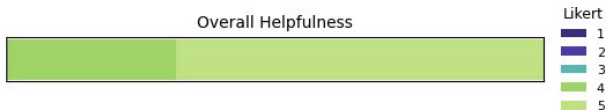

**Figure 5: Overall Helpfulness of Feedback in Improving Writ- ing Skills**

Taken together, these findings highlight both the strengths and current limitations of LLM-based feedback systems. They underscore the importance of designing tools that not only detect surface- level writing features but also attend to the writer's evolving intent, process, and psychological state.

## 5 Future Work

While the present study demonstrates that integrating keylogging and snapshot-based data can enrich the interpretability of LLM-generated feedback, several important avenues remain for further exploration. First, the current dataset—although sufficient for qual- itative thematic saturation—represents an initial cohort. Scaling to a larger and more diverse population will be critical for evaluating the generalizability of these findings and for detecting less frequent feedback patterns.

Second, although this paper focused on triangulating human perceptions of feedback quality, future iterations of the system should include a structured comparison between LLM-generated and human-generated feedback using blind evaluations and rubric-based scoring. This would allow for a more rigorous analysis of alignment, disagreement, and complementarity across modalities. Finally, while this study emphasized cognitive grounding in feedback, students' recurring calls for personalization—including tone sensitivity and genre-awareness—suggest future work should explore integrating user modeling, style preferences, and affective data. This may involve refining prompt-engineering strategies or fine-tuning models on stylistically rich, learner-centered corpora. In sum, the system evaluated here represents a step toward more transparent and process-aware educational AI. As the field moves forward, our findings underscore the importance of designing feed- back systems that are not only linguistically competent but also pedagogically aligned with how learners think, revise, and grow.

## 6 Conclusion

Keylogging and snapshot analysis of the writing process help improve the scoring of LLM feedback for essay writing. Integrating temporal and revision-features greatly boosts the ability of LLM feedback to respond to the writer's changing thoughts.

We aim to expand our efforts by producing a large dataset that provides access to process-traced essay drafts, annotated examples, keylogging details and several forms of feedback (both from humans and AI-based models). Using this dataset will support break- throughs in feedback generation, research on educational NLP and human-AI writing teamwork. Work in the future will look at ways to improve instruction and input methods for LLMs, so they become better and more useful in helping people with writing tasks.